\documentclass[11pt]{article}

\usepackage[utf8]{inputenc}
\usepackage[T1]{fontenc}
\usepackage{amsmath}
\usepackage{amssymb}
\usepackage{amsthm}
\usepackage{booktabs}
\usepackage{graphicx}
\usepackage{hyperref}
\usepackage{microtype}
\usepackage{enumitem}

\IfFileExists{neurips_2025.sty}{
  \usepackage[final,main]{neurips_2025}
}{
  \IfFileExists{neurips_2024.sty}{
    \usepackage[final]{neurips_2024}
  }{
    \usepackage{geometry}
    \geometry{margin=1in}
  }
}

\title{Belief-State RWKV for Reinforcement Learning under Partial Observability}
\author{
  Liu Xiao \\
  \texttt{liu.xiao.in@gmail.com}
}
\date{}

\newtheorem{proposition}{Proposition}
\newtheorem{assumption}{Assumption}

\begin{document}
\maketitle

\begin{abstract}
We propose a stronger formulation of RL on top of RWKV-style recurrent sequence
models, in which the fixed-size recurrent state is explicitly interpreted as a
\emph{belief state} rather than an opaque hidden vector. Instead of conditioning
policy and value on a single summary $h_t$, we maintain a compact
uncertainty-aware state $b_t=(\mu_t,\Sigma_t)$ derived from RWKV-style
recurrent statistics and let control depend on both memory and uncertainty.
This design targets a key weakness of plain fixed-state policies in partially
observed settings: they may store evidence, but not necessarily confidence. We
present the method, a theoretical program, and a pilot RL experiment with
hidden episode-level observation noise together with a test-time noise sweep.
The pilot shows that belief-state policies nearly match the best recurrent
baseline overall while slightly improving return on the hardest
in-distribution regime and under a held-out noise shift. Additional ablations
show that this simple belief readout is currently stronger than two more
structured extensions, namely gated memory control and privileged belief
targets, underscoring the need for richer benchmarks. Code is available at
\url{https://github.com/xiaol/Autoresearch_ideas}.
\end{abstract}

\section{Introduction}
RWKV shows that a recurrent architecture can retain constant-space inference
while still supporting transformer-like parallel training
\cite{peng2023rwkv,peng2024eagle}. This suggests an appealing direction for
reinforcement learning: use an RWKV-style recurrent state as the sole interface
between long-horizon history and decision making. Our earlier formulation
attached policy and value heads directly to a generic linear RNN state $h_t$.
While simple, that view leaves a core
question unresolved: if the agent is uncertain about the latent state of the
environment, where is that uncertainty represented?

That question is especially important because RL generalization often creates
implicit partial observability even when the nominal task description appears
fully observed \cite{ghosh2021epistemicpomdp}. In other words, a recurrent
policy is not only compressing history; it is implicitly constructing belief.

We argue that the next step is to reinterpret the RWKV state as a
\textbf{belief state}. Concretely, instead of a single hidden summary, we maintain
two coupled fixed-size components: a location statistic $\mu_t$ and an
uncertainty statistic $\Sigma_t$. Policy and value are conditioned on both. This
creates a compact interface between memory, uncertainty, and control while
retaining the efficiency of RWKV-style recurrence.

\paragraph{Contributions.}
\begin{itemize}[leftmargin=1.25em]
  \item We introduce a belief-state variant of RL-conditioned RWKV-style models that
  conditions policy and value on $(\mu_t,\Sigma_t)$.
  \item We formalize proposition-level statements around approximate
  sufficiency, stability, and low-rank reward-relevant state structure.
  \item We provide a pilot partially observed RL experiment with hidden
  observation noise, showing promising gains in the hardest and shifted regimes.
  \item We add ablation and calibration diagnostics showing that the plain
  belief-state readout is the strongest simple out-of-distribution variant in
  the current suite.
\end{itemize}

\section{Related Work}
Our work sits at the intersection of RWKV-style recurrent sequence modeling,
recurrent RL under partial observability, and state representation learning.
RWKV reframes the efficient-sequence-model tradeoff by combining parallelizable
training with recurrent inference \cite{peng2023rwkv}, and Eagle/Finch extend
the architecture with matrix-valued states and dynamic recurrence
\cite{peng2024eagle}. In RL, Decision Mamba adapts selective state spaces to
sequence modeling for control \cite{ota2024decisionmamba}, while KalMamba moves
closer to our view by combining probabilistic state-space models with efficient
sequence backbones for RL under uncertainty \cite{becker2024kalmamba}. On the
broader RL side, recent surveys emphasize that representation quality and
uncertainty handling are central to sample efficiency and generalization
\cite{echchahed2025survey}. Generalization-focused analyses further argue that
many practical RL problems should be treated as epistemic POMDPs
\cite{ghosh2021epistemicpomdp}, and recurrent model-free agents remain strong
baselines on such tasks \cite{hausknecht2015drqn,kapturowski2019r2d2,ni2021recurrent}.
Transformer baselines such as GTrXL show that better long-range sequence
processing can matter substantially in RL, but they typically sacrifice the
constant-state deployment story that makes RWKV attractive
\cite{parisotto2019gtrxl}. Our proposal is closer in spirit to predictive-state
approaches that try to expose belief-like recurrent structure directly
\cite{hefny2018rpsp,downey2017psrnn} and to probabilistic latent-state methods
such as DVRL \cite{igl2018dvrl}, but we focus on the control interface of an
RWKV-like backbone itself. This complements benchmark efforts such as POPGym
and newer memory-improvable suites \cite{morad2023popgym,tao2025pobax}, along
with model-based work showing strong efficiency gains from learned world models
\cite{hafner2023dreamerv3,krinner2025accelerating}.

\begin{figure}[t]
  \centering
  \includegraphics[width=\linewidth]{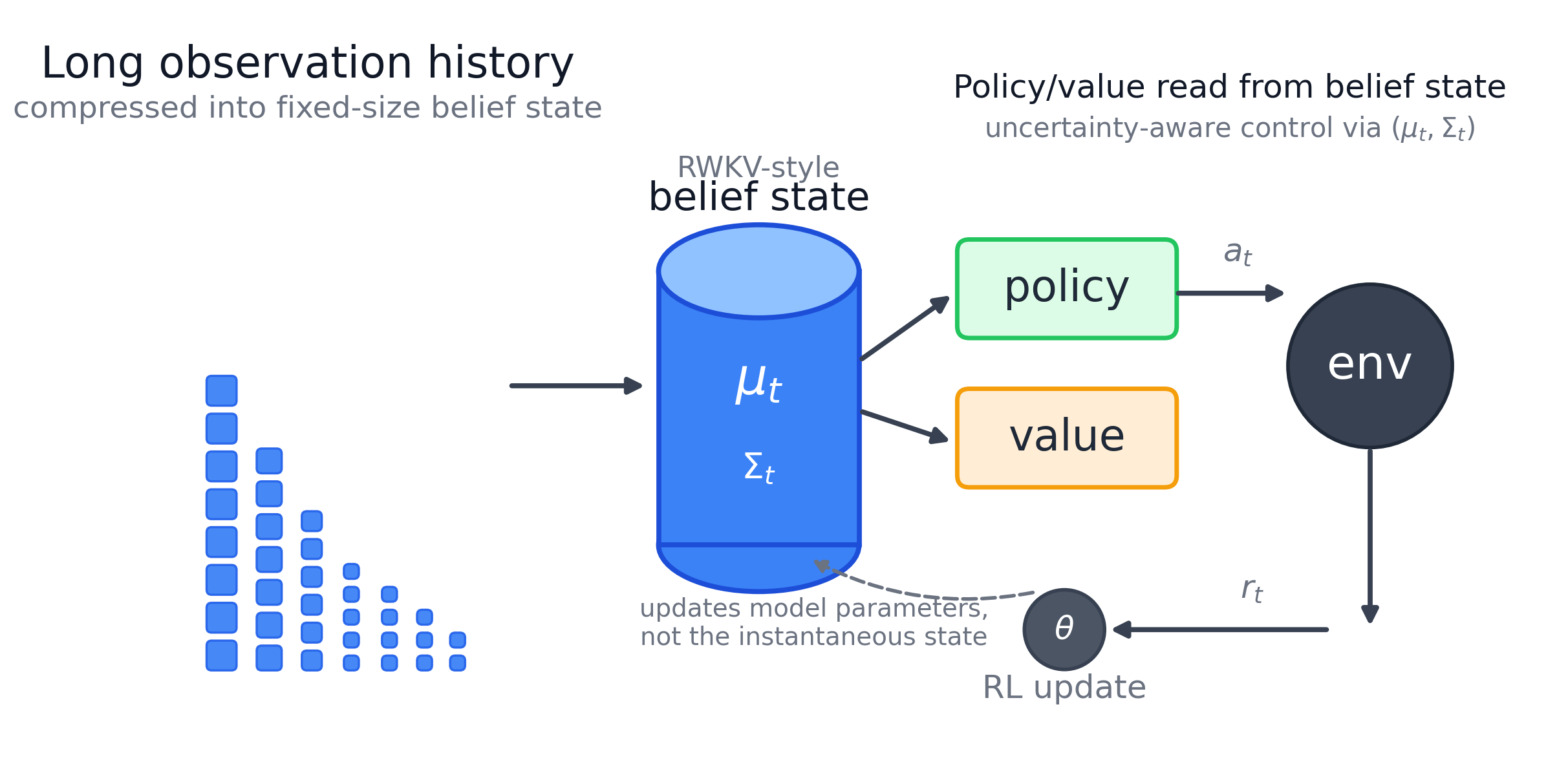}
  \caption{The proposed RWKV-first interface. Long observation history is compressed into a
  fixed-size belief state $(\mu_t,\Sigma_t)$. Policy and value read from this
  uncertainty-aware state, while reward drives parameter updates rather than
  directly defining the instantaneous recurrent state.}
  \label{fig:belief}
\end{figure}

\section{Method}
\subsection{Belief-State RWKV Recurrence}
We replace the opaque hidden state with a structured fixed-size belief state
\begin{equation}
  b_t = (\mu_t,\Sigma_t),
\end{equation}
where $\mu_t$ is a location statistic and $\Sigma_t$ is an uncertainty statistic.
In the simplest case, both are produced by linear recurrent accumulators:
\begin{align}
  s^{(1)}_t &= A_1 s^{(1)}_{t-1} + B_1 x_t, \\
  s^{(2)}_t &= A_2 s^{(2)}_{t-1} + B_2 \phi(x_t),
\end{align}
followed by a deterministic map
\begin{equation}
  \mu_t = f_\mu(s^{(1)}_t), \qquad \Sigma_t = f_\Sigma(s^{(1)}_t,s^{(2)}_t).
\end{equation}
The key design principle is that the state remains fixed-size and recurrent, but
it is now explicitly tasked with representing both \emph{what the agent believes}
and \emph{how certain it is}.

\subsection{RWKV Instantiation}
In a full RWKV instantiation, the belief-state mechanism would sit on top of the
time-mix/channel-mix backbone rather than replacing it. Concretely, one can use
the RWKV recurrent state emitted after the time-mix update as the carrier of
history, then learn lightweight readouts that produce $\mu_t$ and $\Sigma_t$
from that state. The actor and critic then consume $(\mu_t,\Sigma_t)$ instead
of a raw hidden vector. This keeps the computational advantages of RWKV while
making uncertainty explicit at the control interface.

At the block level, let $h_{t-1}$ denote the incoming feature state and let
$s_{t-1}$ denote the RWKV temporal memory. We write the time-mix update
abstractly as
\begin{equation}
  (u_t, s_t) = \mathrm{TimeMix}(x_t, h_{t-1}, s_{t-1}),
\end{equation}
where $u_t$ is the time-mixed feature and $s_t$ is the updated recurrent
memory. A belief readout then branches directly from the temporal pathway:
\begin{equation}
  z_t = \psi(u_t, s_t), \qquad
  \mu_t = W_\mu z_t, \qquad
  \log \Sigma_t = W_\Sigma z_t.
\end{equation}
The feature state passed onward through the backbone is produced by the
channel-mix block,
\begin{equation}
  h_t = \mathrm{ChannelMix}(u_t).
\end{equation}
This placement is intentional: the time-mix state is where RWKV aggregates
history, so it is the natural location from which to expose uncertainty-aware
belief for RL.

\begin{figure}[t]
  \centering
  \includegraphics[width=\linewidth]{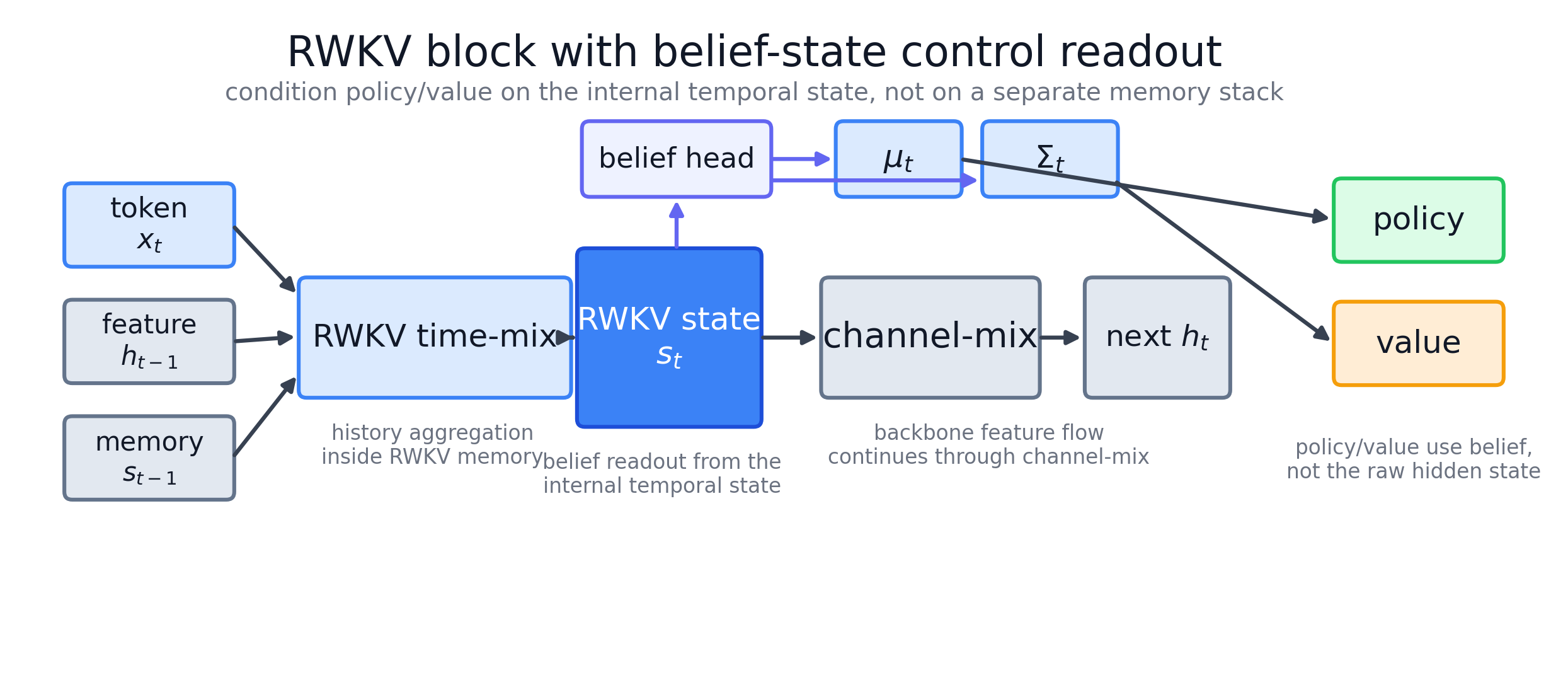}
  \caption{RWKV-specific control interface. The belief readout branches from the
  temporal state produced by RWKV time-mix, then emits $(\mu_t,\Sigma_t)$ for
  policy and value heads. Channel-mix continues to refine the backbone feature
  stream, but the actor-critic reads belief rather than the raw hidden state.}
  \label{fig:rwkv_arch}
\end{figure}

The same design extends naturally to Eagle/Finch-style RWKV variants with
matrix-valued states \cite{peng2024eagle}: the belief readout can be applied to
vector summaries of the matrix state, diagonal uncertainty statistics, or
low-rank projections when a full covariance representation would be too costly.

\subsection{Policy and Value Conditioning}
We define
\begin{equation}
  \pi(a_t \mid \mu_t,\Sigma_t), \qquad V(\mu_t,\Sigma_t),
\end{equation}
and train with a standard actor-critic objective \cite{schulman2017ppo,sutton2018}:
\begin{equation}
  \mathcal{L}
  =
  \mathcal{L}_{\text{policy}}
  + c_v \mathcal{L}_{\text{value}}
  - c_e \mathcal{H}(\pi).
\end{equation}
This formulation is intentionally lightweight: no attention over full history is
required at decision time, and no external memory is assumed beyond the RWKV
state itself.

\subsection{Belief-Conditioned Memory Control}
The belief readout need not be purely passive. RWKV-style recurrence naturally
supports an additional control path in which belief statistics modulate memory
retention itself. Let $s_t^{\text{carry}}$ and $s_t^{\text{write}}$ denote the
carry and write components exposed by the temporal update. We can define
\begin{equation}
  g_t = \sigma\!\left(W_g[\mu_t ; \log \Sigma_t]\right), \qquad
  s_t^{\text{ctrl}} = g_t \odot s_t^{\text{carry}} + (1-g_t) \odot s_t^{\text{write}}.
\end{equation}
High uncertainty can therefore increase the effective write rate, while low
uncertainty favors retention of stable evidence. We do not activate this
feedback path in the pilot experiment, but it is a natural RWKV-specific
extension because the temporal state already exposes a controllable
carry-versus-write decomposition.

\subsection{Low-Rank Belief Adapters}
One natural extension is to augment the belief state with a low-rank task
adapter
\begin{equation}
  \tilde{b}_t = b_t + U g(b_t),
\end{equation}
where $U$ is low rank. This lets the policy specialize to reward-relevant
subspaces without replacing the recurrent backbone. We view this as especially
promising for RWKV variants where only a small portion of the recurrent state
may be reward-relevant at any given time.

\subsection{Privileged Belief Supervision}
In simulator-based RL benchmarks, one often has access during training to
latent variables that are hidden at test time. This suggests an auxiliary
objective for belief-state RWKV:
\begin{equation}
  \mathcal{L}_{\text{belief}}
  =
  \beta_\mu \|\mu_t - \mu_t^\star\|_2^2
  +
  \beta_\Sigma \|\log \Sigma_t - \log \Sigma_t^\star\|_2^2,
\end{equation}
where $(\mu_t^\star,\Sigma_t^\star)$ are posterior moments or simulator-derived
belief targets. This keeps inference fully recurrent while using privileged
signal only as a training regularizer, in a similar spirit to guided learning
under partial observability \cite{li2025gpo} and recent privileged-information
world-model approaches \cite{huang2025pigdreamer}. We view this as a practical
path toward more interpretable internal RWKV states.

\section{Theory}
We next state three proposition-level results that make the research program
more concrete. We present proof sketches rather than full formal derivations.

\begin{assumption}[Approximate predictive sufficiency]
Let $H_t$ denote the interaction history and $S_t$ the latent POMDP state.
There exists a belief encoder $\psi$ and a decoder $q$ such that for all
histories,
\begin{equation}
  \mathrm{TV}\!\left(p(S_t \mid H_t), q(S_t \mid \psi(H_t))\right) \le \varepsilon.
\end{equation}
\end{assumption}

\begin{proposition}[Approximate sufficiency bound]
Under the assumption above, bounded rewards $|r| \le 1$, and a transition kernel
whose one-step predictive error is controlled by the same total-variation gap,
the value difference between the optimal history-dependent policy and the
optimal policy acting only on $b_t=\psi(H_t)$ satisfies
\begin{equation}
  \sup_{H_t}
  \left|
    V^\star(H_t) - V^\star_\psi(\psi(H_t))
  \right|
  \le
  \frac{2\varepsilon}{(1-\gamma)^2}.
\end{equation}
\end{proposition}

\paragraph{Proof sketch.}
The decoder $q$ induces an approximate belief-MDP whose one-step reward and
transition errors are both $O(\varepsilon)$. A standard simulation-lemma style
argument then converts this local model mismatch into a discounted value gap.
The extra factor of $(1-\gamma)^{-1}$ beyond Bellman contraction comes from
propagating one-step belief error through future occupancy.

\begin{assumption}[Stable linear recurrence]
Each linear recurrent block satisfies $\|A_i\|_2 \le \rho < 1$, the input maps
are bounded, and the readout maps $f_\mu,f_\Sigma$ are Lipschitz on bounded
sets.
\end{assumption}

\begin{proposition}[Bounded belief-state trajectory]
Under the stability assumption, there exists a constant $C$ depending on input
scale and the Lipschitz constants of the readouts such that
\begin{equation}
  \sup_t \|b_t\|_2 \le \frac{C}{1-\rho}.
\end{equation}
\end{proposition}

\paragraph{Proof sketch.}
Repeated application of the linear recurrence yields a geometric series in
$\rho$. Because the readout maps are Lipschitz, bounded recurrent statistics
imply bounded $(\mu_t,\Sigma_t)$. This proposition is intentionally simple, but
it captures the core reason RWKV-style fixed-state control is easier to
stabilize than arbitrary nonlinear recurrent belief updates.

\begin{assumption}[Low-rank reward relevance]
There exists a projection $P_r$ onto an $r$-dimensional subspace of the belief
state such that for every action $a$,
\begin{equation}
  \left|Q^\pi(b,a) - Q^\pi(P_r b, a)\right| \le \delta_r.
\end{equation}
\end{assumption}

\begin{proposition}[Low-rank adapter approximation]
If the policy and value heads act on a low-rank adapted belief
$\tilde{b}=P_r b + U g(b)$ with $U \in \mathbb{R}^{d \times r}$, then the
suboptimality induced by restricting control to the reward-relevant rank-$r$
subspace is at most
\begin{equation}
  O\!\left(\frac{\delta_r}{1-\gamma}\right).
\end{equation}
\end{proposition}

\paragraph{Proof sketch.}
The assumption states that truncating belief state outside the reward-relevant
subspace perturbs action values by at most $\delta_r$. Applying the performance
difference lemma bounds policy loss in terms of that perturbation. This is the
formal justification for using low-rank belief adapters instead of full-rank
task-specific recurrent controllers.

\section{Pilot Experiment}
\subsection{Environment}
We study a simple partially observed \emph{stop-or-guess} environment. Each
episode samples a hidden label $z \in \{-1,+1\}$ and a hidden episode-level
observation noise $\sigma \sim \mathcal{U}(0.3,1.2)$. At step $t$, the agent
observes
\begin{equation}
  x_t = z + \epsilon_t, \qquad \epsilon_t \sim \mathcal{N}(0,\sigma^2).
\end{equation}
The agent can either wait (small penalty) or commit to one of two guesses. A
correct guess yields $+1$, an incorrect guess yields $-1$, and waiting costs
$0.05$. Because $\sigma$ is hidden and varies by episode, a good policy must
reason jointly about accumulated evidence and uncertainty.

\subsection{Models}
We compare three actor-critic policies:
\begin{itemize}[leftmargin=1.25em]
  \item \textbf{MLP (memoryless):} acts from the current observation only.
  \item \textbf{RWKV-style summary state:} conditions on a recurrent evidence
  summary.
  \item \textbf{Belief-state RWKV-style:} conditions on running mean and running
  uncertainty statistics derived from RWKV-style recurrent accumulators.
\end{itemize}
All models are trained for 1600 optimization steps with batch size 256 and
sequence length 10 over 3 random seeds. We emphasize that this pilot is a
RWKV-style control abstraction rather than a full pretrained RWKV backbone; the
purpose is to isolate the effect of explicit belief-state conditioning.

\begin{table}[t]
\centering
\begin{tabular}{l c c c}
\toprule
Model & Mean return & Hard return & Very-hard return \\
\midrule
MLP (memoryless) & 0.909 $\pm$ 0.006 & 0.836 $\pm$ 0.017 & 0.809 $\pm$ 0.018 \\
RWKV-style summary state & \textbf{0.924 $\pm$ 0.008} & 0.861 $\pm$ 0.016 & 0.820 $\pm$ 0.021 \\
Belief-state RWKV-style & 0.919 $\pm$ 0.013 & \textbf{0.862 $\pm$ 0.025} & \textbf{0.822 $\pm$ 0.030} \\
\bottomrule
\end{tabular}
\caption{Pilot results on the hidden-noise stop-or-guess task. The summary-state
model performs best overall, but the belief-state model slightly improves return
in the hardest noise regimes, suggesting that explicit uncertainty can help when
partial observability is most severe.}
\label{tab:pilot}
\end{table}

\paragraph{Shift protocol.}
To test robustness under distribution shift, we also train on an easier noise
range $\sigma \sim \mathcal{U}(0.3, 1.2)$ and evaluate without retraining on a
harder held-out range $\sigma \sim \mathcal{U}(1.2, 1.8)$.

\begin{table}[t]
\centering
\begin{tabular}{l c c}
\toprule
Model & OOD mean return & OOD very-hard return \\
\midrule
MLP (memoryless) & 0.630 $\pm$ 0.008 & 0.630 $\pm$ 0.008 \\
RWKV-style summary state & 0.643 $\pm$ 0.011 & 0.643 $\pm$ 0.011 \\
Belief-state RWKV-style & \textbf{0.650 $\pm$ 0.015} & \textbf{0.650 $\pm$ 0.015} \\
\bottomrule
\end{tabular}
\caption{Held-out noise-shift evaluation. The belief-state model performs best
when trained on easier episodes and tested on a strictly harder observation
noise regime, supporting the claim that explicit uncertainty is especially
useful under distribution shift.}
\label{tab:ood}
\end{table}

\paragraph{Interpretation.}
The pilot result is intentionally modest. We do not yet claim that belief-state
conditioning dominates summary-state conditioning across all settings. Instead,
the result supports a more specific claim: when observation noise is hidden and
varies across episodes, explicit uncertainty tracking is competitive overall,
most helpful in the hardest subset, and slightly more robust under shift. This
is exactly the regime where plain recurrent summaries are least interpretable.

\paragraph{Ablations and calibration.}
We also test two method extensions motivated by our RWKV formulation. The first
replaces fixed accumulation with an adaptive write gate, yielding a
\emph{belief-state + gated memory} policy. The second adds an auxiliary loss
toward the simulator posterior moments over the hidden label, yielding a
\emph{belief-state + privileged targets} policy. In addition to return, we
measure decision-time expected calibration error (ECE) by renormalizing the two
guess logits into a binary posterior over $z$ at the step where the policy
commits.

\begin{table}[t]
\centering
\begin{tabular}{l c c c c}
\toprule
Model & ID return & OOD return & ID ECE & OOD ECE \\
\midrule
RWKV-style summary state & \textbf{0.924} & 0.643 & 0.024 & 0.164 \\
Belief-state RWKV-style & 0.919 & \textbf{0.650} & 0.025 & \textbf{0.160} \\
Belief-state + gated memory & 0.917 & 0.643 & \textbf{0.024} & 0.162 \\
Belief-state + privileged targets & 0.921 & 0.634 & 0.027 & 0.170 \\
\bottomrule
\end{tabular}
\caption{Ablation and calibration summary. The plain summary-state RWKV baseline
remains strongest on in-distribution return, but the plain belief-state RWKV
variant is still the best simple out-of-distribution choice and has the best
held-out calibration. Extra structure does not automatically help on this toy
task: gated memory is competitive but not better, while privileged targets
reduce robustness under shift.}
\label{tab:ablation}
\end{table}

The ablation result is useful precisely because it is not uniformly positive.
The basic belief-state readout remains the strongest OOD variant in this suite.
Adaptive memory gating slightly improves in-distribution calibration without
improving held-out return, while privileged belief targets appear to accelerate
decision making but overspecialize to the training regime. Taken together, the
result suggests that belief \emph{exposure} is already helpful, whereas belief
\emph{control} and belief \emph{supervision} likely need richer benchmarks than
the current stop-or-guess task to show their full value.

\begin{figure}[t]
  \centering
  \includegraphics[width=\linewidth]{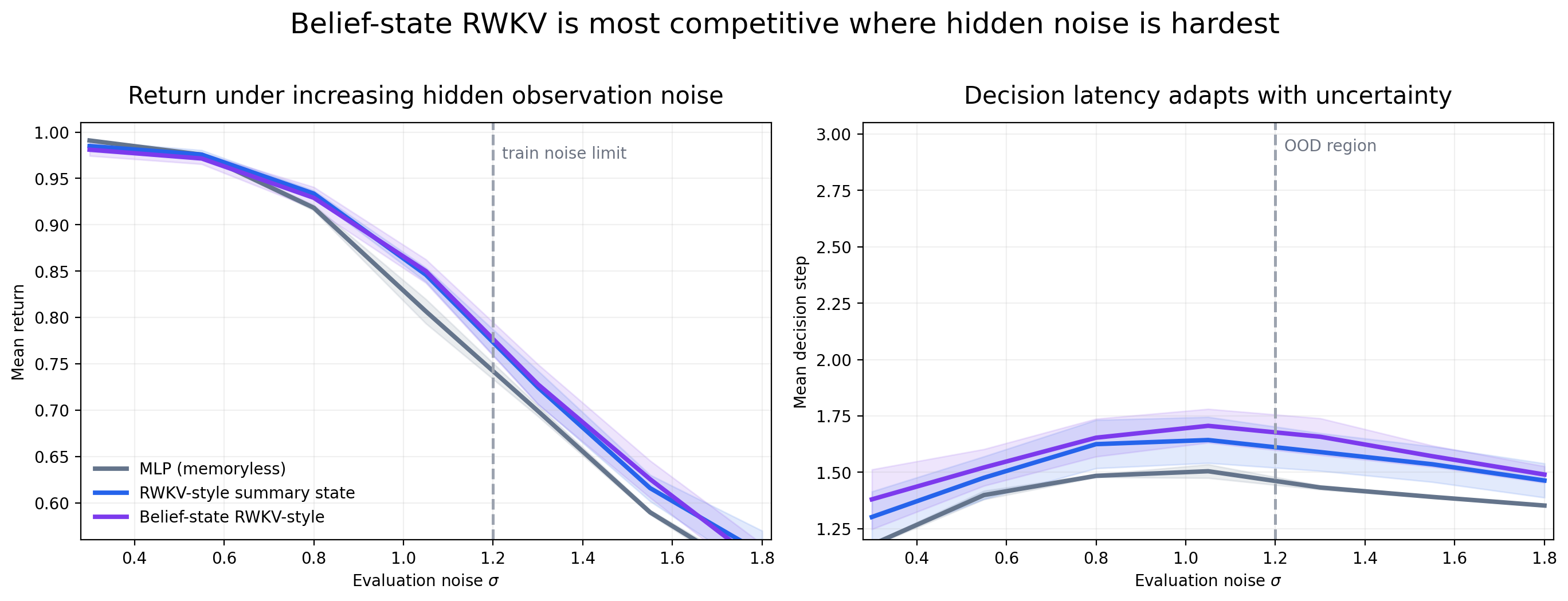}
  \caption{Robustness sweep across fixed evaluation noise levels. The summary-state
  RWKV baseline remains strongest on the easy side of the training range, but
  the belief-state variant degrades more slowly once hidden noise approaches and
  exceeds the train-time ceiling. The right panel shows that this comes with a
  smoother increase in decision latency, consistent with uncertainty-aware
  waiting rather than indiscriminate hesitation.}
  \label{fig:sweep}
\end{figure}

The sigma sweep in Figure~\ref{fig:sweep} makes the tradeoff more visible than
the aggregate tables alone. On easy episodes, the summary-state recurrent
baseline commits slightly earlier and enjoys the best mean return. As hidden
noise rises, however, the belief-state policy adapts its stopping behavior more
smoothly and is strongest across much of the hard and out-of-distribution
regime.

\section{Main Experimental Agenda}
The pilot is only a first step. A full evaluation should test three hypotheses:
\begin{enumerate}[leftmargin=1.25em]
  \item \textbf{Partial-observability scaling:} belief-state models should widen
  their advantage as the gap between observable evidence and latent state grows.
  \item \textbf{Distribution shift:} uncertainty-aware states should improve
  robustness when observation noise, distractors, or horizon length shift at
  test time.
  \item \textbf{Efficiency versus attention:} fixed-size belief states should
  preserve much of the benefit of recurrent efficiency while closing part of the
  decision-quality gap to heavier attention-based or world-model baselines.
\end{enumerate}

Concretely, we see three immediate benchmark families:
\begin{itemize}[leftmargin=1.25em]
  \item Synthetic POMDPs with hidden volatility, delayed rewards, and decision
  deadlines.
  \item Memory-improvable benchmark suites where confidence calibration matters
  as much as latent-state estimation \cite{morad2023popgym,tao2025pobax}.
  \item Long-horizon sequence-control settings for full RWKV backbones,
  including Eagle/Finch-style state variants, with comparisons against strong
  recurrent and world-model baselines \cite{hausknecht2015drqn,kapturowski2019r2d2,hafner2023dreamerv3}.
\end{itemize}

The current shift result already offers a small but concrete signal in favor of
the second hypothesis: explicit uncertainty did not maximize the in-distribution
average return, but it did produce the strongest held-out performance when test
noise exceeded the training range.

\section{Discussion}
The main conceptual advantage of the RWKV belief-state view is not just
performance; it is \emph{interface clarity}. A plain hidden state can in principle encode
everything, but it gives the researcher little leverage over what is stored.
Belief-state factorization adds structure without giving up recurrence. It opens
doors to uncertainty-aware policy improvement, interpretable state diagnostics,
and theorem-friendly assumptions about sufficiency and stability.

At the same time, the pilot shows that this is not a free lunch. Explicit
uncertainty channels can help in hard regimes without automatically improving the
mean case. The design problem is therefore not whether to add uncertainty, but
how to represent it compactly and use it selectively. Our own results reinforce
that point: the plain summary-state baseline remains strongest on average
in-distribution, while the belief-state variant becomes most attractive in the
tails and under held-out shift. The most promising next step is therefore not
simply ``more uncertainty'', but tighter integration between belief and RWKV
memory management itself. Our ablations show that uncertainty-gated retention
and privileged belief supervision are reasonable next ideas, but not yet solved:
they need stronger partial-observability benchmarks before they can be judged
fairly.

\section{Conclusion}
We propose belief-state RWKV as a stronger successor to plain RL-conditioned
hidden-state control. The idea is simple: keep the fixed-size RWKV recurrent
memory, but structure it as uncertainty-aware belief rather than an opaque
vector. The resulting interface is compact, recurrent, and better aligned with
partial observability. Our pilot experiment provides early evidence that this
direction is most useful precisely where uncertainty matters most, and the
theoretical program suggests several clear next steps for a full RWKV-centered
paper. The new ablation result sharpens that conclusion: a simple belief
readout already helps under shift, while more structured variants such as gated
memory control and privileged belief targets still need stronger benchmarks to
prove their worth.

\bibliographystyle{plainnat}
\bibliography{refs}

@article{peng2023rwkv,
  title={RWKV: Reinventing RNNs for the Transformer Era},
  author={Peng, Bo and Alcaide, Eric and Anthony, Quentin and Albalak, Alon and Arcadinho, Samuel and Biderman, Stella and Cao, Huanqi and Cheng, Xin and Chung, Michael and Grella, Matteo and others},
  journal={arXiv preprint arXiv:2305.13048},
  year={2023}
}

@article{peng2024eagle,
  title={Eagle and Finch: RWKV with Matrix-Valued States and Dynamic Recurrence},
  author={Peng, Bo and Goldstein, Daniel and Anthony, Quentin and Albalak, Alon and Alcaide, Eric and Biderman, Stella and Cheah, Eugene and Du, Xingjian and Ferdinan, Teddy and Hou, Haowen and others},
  journal={arXiv preprint arXiv:2404.05892},
  year={2024}
}

@article{ota2024decisionmamba,
  title={Decision Mamba: Reinforcement Learning via Sequence Modeling with Selective State Spaces},
  author={Ota, Toshihiro},
  journal={arXiv preprint arXiv:2403.19925},
  year={2024}
}

@article{ghosh2021epistemicpomdp,
  title={Why Generalization in RL is Difficult: Epistemic POMDPs and Implicit Partial Observability},
  author={Ghosh, Dibya and Rahme, Jad and Kumar, Aviral and Zhang, Amy and Adams, Ryan P. and Levine, Sergey},
  journal={arXiv preprint arXiv:2107.06277},
  year={2021}
}

@article{hausknecht2015drqn,
  title={Deep Recurrent Q-Learning for Partially Observable MDPs},
  author={Hausknecht, Matthew and Stone, Peter},
  journal={arXiv preprint arXiv:1507.06527},
  year={2015}
}

@inproceedings{kapturowski2019r2d2,
  title={Recurrent Experience Replay in Distributed Reinforcement Learning},
  author={Kapturowski, Steven and Ostrovski, Georg and Quan, John and Munos, R{\'e}mi and Dabney, Will},
  booktitle={International Conference on Learning Representations},
  year={2019},
  url={https://openreview.net/forum?id=r1lyTjAqYX}
}

@article{ni2021recurrent,
  title={Recurrent Model-Free RL Can Be a Strong Baseline for Many POMDPs},
  author={Ni, Tianwei and Eysenbach, Benjamin and Salakhutdinov, Ruslan},
  journal={arXiv preprint arXiv:2110.05038},
  year={2021}
}

@article{hefny2018rpsp,
  title={Recurrent Predictive State Policy Networks},
  author={Hefny, Ahmed and Marinho, Zita and Sun, Wen and Srinivasa, Siddhartha and Gordon, Geoffrey},
  journal={arXiv preprint arXiv:1803.01489},
  year={2018}
}

@article{downey2017psrnn,
  title={Predictive State Recurrent Neural Networks},
  author={Downey, Carlton and Hefny, Ahmed and Li, Boyue and Boots, Byron and Gordon, Geoffrey},
  journal={arXiv preprint arXiv:1705.09353},
  year={2017}
}

@article{igl2018dvrl,
  title={Deep Variational Reinforcement Learning for POMDPs},
  author={Igl, Maximilian and Zintgraf, Luisa and Le, Tuan Anh and Wood, Frank and Whiteson, Shimon},
  journal={arXiv preprint arXiv:1806.02426},
  year={2018}
}

@article{parisotto2019gtrxl,
  title={Stabilizing Transformers for Reinforcement Learning},
  author={Parisotto, Emilio and Song, H. Francis and Rae, Jack W. and Pascanu, Razvan and Gulcehre, Caglar and Jayakumar, Siddhant M. and Jaderberg, Max and Kaufman, Raphael Lopez and Clark, Aidan and Noury, Seb and Botvinick, Matthew M. and Heess, Nicolas and Hadsell, Raia},
  journal={arXiv preprint arXiv:1910.06764},
  year={2019}
}

@article{becker2024kalmamba,
  title={KalMamba: Towards Efficient Probabilistic State Space Models for RL under Uncertainty},
  author={Becker, Philipp and Freymuth, Niklas and Neumann, Gerhard},
  journal={arXiv preprint arXiv:2406.15131},
  year={2024}
}

@article{li2025gpo,
  title={Guided Policy Optimization under Partial Observability},
  author={Li, Yueheng and Xie, Guangming and Lu, Zongqing},
  journal={arXiv preprint arXiv:2505.15418},
  year={2025}
}

@article{krinner2025accelerating,
  title={Accelerating Model-Based Reinforcement Learning with State-Space World Models},
  author={Krinner, Maria and Aljalbout, Elie and Romero, Angel and Scaramuzza, Davide},
  journal={arXiv preprint arXiv:2502.20168},
  year={2025}
}

@article{hafner2023dreamerv3,
  title={Mastering Diverse Domains through World Models},
  author={Hafner, Danijar and Pasukonis, Jurgis and Ba, Jimmy and Lillicrap, Timothy},
  journal={arXiv preprint arXiv:2301.04104},
  year={2023}
}

@article{echchahed2025survey,
  title={A Survey of State Representation Learning for Deep Reinforcement Learning},
  author={Echchahed, Ayoub and Castro, Pablo Samuel},
  journal={arXiv preprint arXiv:2506.17518},
  year={2025}
}

@article{morad2023popgym,
  title={POPGym: Benchmarking Partially Observable Reinforcement Learning},
  author={Morad, Steven and Kortvelesy, Ryan and Bettini, Matteo and Liwicki, Stephan and Prorok, Amanda},
  journal={arXiv preprint arXiv:2303.01859},
  year={2023}
}

@article{tao2025pobax,
  title={Benchmarking Partial Observability in Reinforcement Learning with a Suite of Memory-Improvable Domains},
  author={Tao, Ruo Yu and Guo, Kaicheng and Allen, Cameron and Konidaris, George},
  journal={arXiv preprint arXiv:2508.00046},
  year={2025}
}

@article{huang2025pigdreamer,
  title={PIGDreamer: Privileged Information Guided World Models for Safe Partially Observable Reinforcement Learning},
  author={Huang, Dongchi and Wang, Jiaqi and Li, Yang and Xia, Chunhe and Zhang, Tianle and Zhang, Kaige},
  journal={arXiv preprint arXiv:2508.02159},
  year={2025}
}

@article{schulman2017ppo,
  title={Proximal Policy Optimization Algorithms},
  author={Schulman, John and Wolski, Filip and Dhariwal, Prafulla and Radford, Alec and Klimov, Oleg},
  journal={arXiv preprint arXiv:1707.06347},
  year={2017}
}

@book{sutton2018,
  title={Reinforcement Learning: An Introduction},
  author={Sutton, Richard S. and Barto, Andrew G.},
  year={2018},
  edition={2},
  publisher={MIT Press}
}
\end{document}